\documentclass[conference]{IEEEtran}
\IEEEoverridecommandlockouts

\usepackage{cite}
\usepackage{amsmath,amssymb,amsfonts}
\usepackage{algorithmic}
\usepackage{graphicx}
\usepackage{textcomp}
\usepackage{xcolor}
\usepackage[]{algorithm2e}
\usepackage{graphicx}  %Required
\usepackage{booktabs}       % professional-quality tables
\usepackage{subcaption}
\usepackage{amsmath}
\def\BibTeX{{\rm B\kern-.05em{\sc i\kern-.025em b}\kern-.08em
		T\kern-.1667em\lower.7ex\hbox{E}\kern-.125emX}}
\begin{document}
	
	\title{Local Critic Training of Deep Neural Networks}
	
	\author{\IEEEauthorblockN{Hojung Lee}
		\IEEEauthorblockA{\textit{School of Integrated Technology, Yonsei University, Korea} \\
			hjlee92@yonsei.ac.kr}
		\and
		\IEEEauthorblockN{Jong-Seok Lee}
		\IEEEauthorblockA{\textit{School of Integrated Technology, Yonsei University, Korea} \\
			jong-seok.lee@yonsei.ac.kr}
	}
	
	\maketitle

	\begin{abstract}
		This paper proposes a novel approach to train deep neural networks by unlocking the layer-wise dependency of backpropagation training.
		The approach employs additional modules called local critic networks besides the main network model to be trained, which are used to obtain error gradients without complete feedforward and backward propagation processes.
		We propose a cascaded learning strategy for these local networks. 
		In addition, the approach is also useful from multi-model perspectives, including structural optimization of neural networks, computationally efficient progressive inference, and ensemble classification for performance improvement. 
		Experimental results show the effectiveness of the proposed approach and suggest guidelines for determining appropriate algorithm parameters.
	\end{abstract}
	
	\section{Introduction}
	
	\label{submission}
	In recent days, deep learning has been remarkably advanced and successfully applied in numerous fields \cite{LeCun15}.
	A key mechanism behind the success of deep neural networks is that they are capable of extracting useful information progressively through their layered structures.
	It is an increasing trend that more and more complex deep neural network structures are developed in order to solve challenging real-world problems, e.g., \cite{he16identity}.
	Training of deep neural networks is based on backpropagation in most cases, which basically works in a sequential and synchronous manner. 
	During the feedforward pass, the input data is processed through the hidden layers to produce the network output; during the feedback pass, the error gradient is propagated back through the layers to update each layer's weight parameters. Therefore, training of each layer has dependency on all the other layers, which causes the issue of \textit{locking} \cite{Jaderberg17}. This is undesirable in some cases, e.g., a system consisting of several interacting models, a model distributed across multiple computing nodes, etc.
	
	% DNI
	There have been attempts to remove the locking constraint. In \cite{Carreira14}, the method of auxiliary coordinates (MAC) is proposed. It replaces the original loss minimization problem with an equality-constrained optimization problem by introducing an auxiliary variable for each data and each hidden unit. Then, solving the problem is formulated as iteratively solving several sub-problems independently. A similar approach using the alternating direction method of multipliers (ADMM) is proposed in \cite{Taylor16}. It also employs an equality-constrained optimization but with different auxiliary variables, so that resulting sub-problems have closed form solutions. However, these methods are not scalable to deep learning architectures such as convolutional neural networks (CNNs). 
	
	The method proposed in \cite{Jaderberg17}, called decoupled neural interface (DNI), directly synthesizes estimated error gradients, called synthetic gradients, using an additional small neural network for training a layer's weight parameters. 
	As long as the synthetic gradients are close to the actual backpropagated gradients, each layer does not need to wait until the error at the output layer is backpropagated through the preceding layers, which allows independent training of each layer.
	However, this method suffers from performance degradation when compared to regular backpropagation \cite{Czarnecki17}.
	% sobolev
	The idea of having additional modules supporting the layers of the main model is also adopted in \cite{Czarnecki17}, where the additional modules are trained to approximate the main model's outputs instead of error gradients.
	Due to this, however, the method does not resolve the issue of update locking, and in fact, the work does not intend to design a non-sequential learning algorithm.
	
	In this paper, we propose a novel approach for non-sequential learning, called \textit{local critic training}.
	The key idea is that additional modules besides the main neural network model are employed, which we call \textit{local critics}, in order to indirectly deliver error gradients to the main model for training without backpropagation.
	In other words, a local critic located at a certain layer group is trained in such a way that the derivative of its output serves as the error gradient for training of the corresponding layers' weight parameters.
	Thus, the error gradient does not need to be backpropagated, and the feedforward operations and gradient-descent learning can be performed independently.
	%We show that the proposed structure combining a main network and local critics is a universal approximator.
	Through extensive experiments, we examine the influences of the network structure, update frequency, and total number of local critics, which provide not only insight into operation characteristics but also guidelines for performance optimization of the proposed method.
	
	In addition to the capability of implementing training without locking, the proposed approach can be exploited for additional important applications.
	First, we show that applying the proposed method automatically performs structural optimization of neural networks for a given problem, which has been a challenging issue in the machine learning field.
	Second, a progressive inference algorithm using the network trained with the proposed method is presented, which can adaptively reduce the computational complexity during the inference process (i.e., test phase) depending on the given data.
	Third, the network trained by the proposed method naturally enables ensemble inference that can improve the classification performance.
	
	%%%%%%%%%%%%%%%%%%%%%%%%%%%%%%%%%%%%%%%%%%%%%%
	\section{Proposed Approach}
	\label{sec:method}
	\begin{figure*}[t]
		\centering
		{\includegraphics[width=5in]{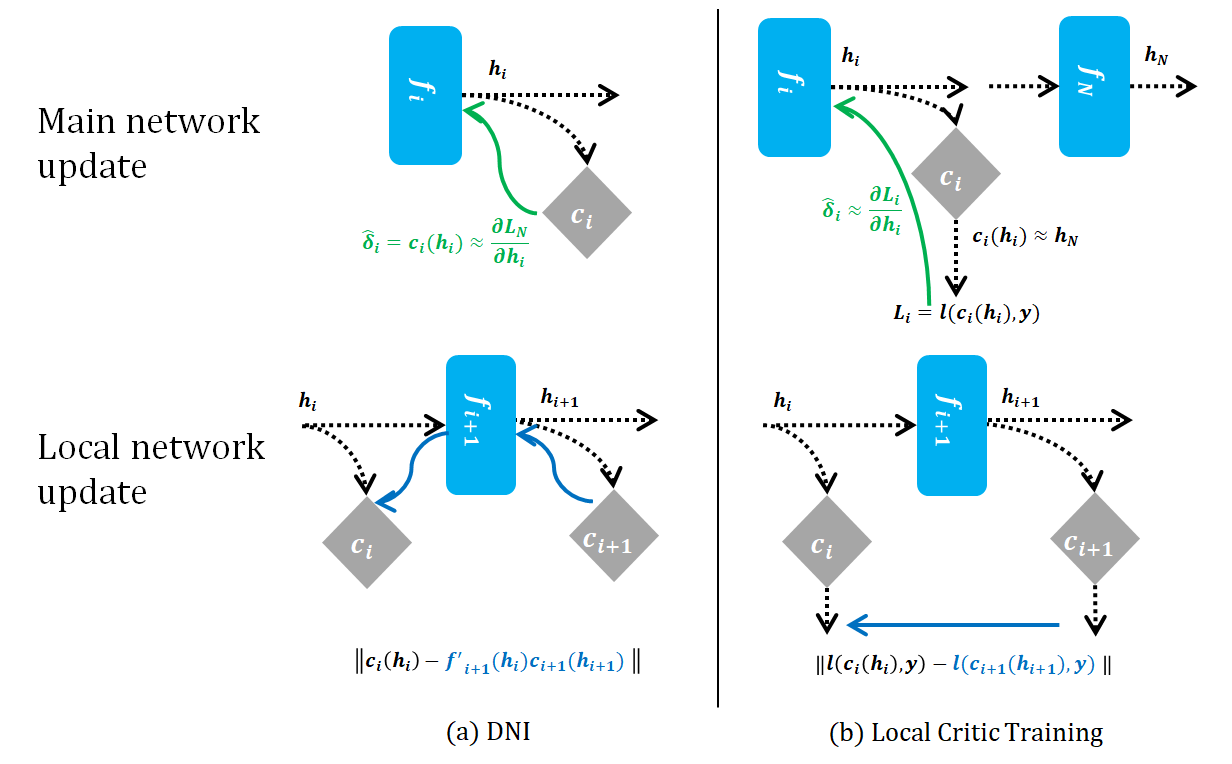}}
		\caption{Learning processes of DNI \cite{Jaderberg17} and the proposed local critic training. The black, green, and blue arrows indicate feedforward passes, an error gradient flow, and loss comparison, respectively.}
		\label{fig:comparison}
	\end{figure*}
	\subsection{Local Critic Training}
	The basic idea of the proposed approach is to introduce additional local networks, which we call local critics, besides the main network model, so that they eventually provide estimates of the output of the main network.
	Each local critic network can serve a group of layers of the main model by being attached to the last layer of the group.
	The proposed architecture is illustrated in Figure \ref{fig:comparison}, where $f_i$ is the $i$th layer group (containing one or more layers), $h_i$ is the output of $f_i$, and $h_N$ is the final output of the main model having $N$ layer groups:
	\begin{equation}
	\label{eq0}
	h_i = f_i(h_{i-1})
	\end{equation}
	$c_i$ is the local critic network for $f_i$, which is expected to approximate $h_N$ based on $h_i$, i.e.,
	\begin{equation}
	\label{eq1}
	c_{i}(h_{i}) \approx h_{N}
	%l(c_{i}(h_{i}|\theta_i), y) \approx l(h_{N}, y) 
	\end{equation}
	Then, this can be used to approximate the loss function of the main network, $L_N=l(h_N,y)$, which is used to train $f_i$, by %the training loss for $f_i$ can be approximated as
	\begin{equation}
	\label{eq2}
	L_i = l(c_i(h_i), y) 
	\end{equation}
	for $i=1,...,N-1$, i.e.,
	\begin{equation}
	\label{eq100}
	L_i \approx L_N
	\end{equation}
	where $y$ is the training target and $l$ is the loss function such as cross-entropy or mean-squared error.
	%\begin{equation}
	%\label{eq2}
	%L_{i} = l(c_{i}(h_{i}), y)
	%\end{equation}
	%where $y$ is the training target and $l$ is the loss function such as cross-entropy or mean-squared error.
	Then, the error gradient for training $f_i$ is obtained by differentiating $L_i$ with respect to $h_i$, i.e.,
	\begin{equation}
	\label{eq3}
	\delta_{i} = \frac{\partial L_{i}}{\partial h_{i}}
	\end{equation}
	which can be used to train the weight parameters of $f_i$, denoted by $\theta_i$, via a gradient-descent rule:
	\begin{equation}
	\theta_i \leftarrow \theta_i - \eta ~\delta_i ~\frac{\partial h_i}{\partial \theta_i}
	\end{equation}
	where $\eta$ is a learning rate. Note that the final layer group $h_N$ does not require a local critic network and can be trained using the regular backproagation because the final output of the main network is directly available.
	Therefore, the update of $f_i$ does not need to wait until its output $h_i$ propagates till the end of the main network and the error gradient is backpropagated; it can be performed when the operations from (\ref{eq1}) to (\ref{eq3}) are done.
	For $c_i$, we usually use a simple model so that the operations through $c_i$ are simpler than those through $f_{i+1}$ till $f_N$.
	
	While the dependency of $f_i$ on $f_j$ ($j>i$) during training is resolved in this way, there still exists the dependency of $c_i$ on $f_j$ ($j>i$), because training $c_i$ requires its ideal target, i.e., $h_N$, which is available from $f_N$ only after the feedforward pass is complete.
	In order to resolve this problem, we use an indirect, cascaded approach, where $c_i$ is trained so that its training loss targets the training loss for $c_{i+1}$\footnote{We found that this is more effective than directly forcing $c_i$ to approximate $c_{i+1}$ using $L_{c_i}=l(c_i(h_i), c_{i+1}(h_{i+1}))$.}:
	\begin{equation}
	\label{eq4}
	L_{c_i}=l(L_i, L_{i+1})
	\end{equation}
	In other words, training of $c_i$ can be performed once the loss for $c_{i+1}$ is available.
	
	Figure \ref{fig:comparison} compares the proposed architecture with the existing DNI approach that also employs local networks besides the main network to resolve the issue of locking \cite{Jaderberg17}.
	In DNI, the local network $c_i$ directly estimates the error gradient, i.e.,
	\begin{equation}
	c_i(h_i) \approx \frac{\partial L_N}{\partial h_i}
	\end{equation}
	so that each layer group of the main model can be updated without waiting for the forward and backward propagations in the subsequent layers.
	And, to update $c_i$, the error gradient for $f_{i+1}$ estimated by $c_{i+1}$ is backpropagated through $f_{i+1}$ and is used as the (estimated) target for $c_i$.
	Therefore, all the necessary computations in the forward and backward passes can be locally confined. %, which enables non-sequential training.
	The performance of the two methods will be compared in Section \ref{sec:exp}.
	
	%%%%%%%%%%%%%%%%%%%%%%%%
	\subsection{Structural optimization}
	
	In many cases, determining an appropriate structure of neural networks for a given problem is not straightforward.
	This is usually done through trial-and-error, which is extremely time-consuming.
	There have been studies to automate the structural optimization process \cite{Cortes17,feng15,kwon97,reed93}, but this issue still remains very challenging.
	
	In deep learning, the problem of structural optimization is even more critical.
	Large-sized networks may easily show overfitting.
	Even if large networks may produce high accuracy, they take significantly large amounts of memory and computation, which is undesirable especially for resource-constrained cases such as embedded and mobile systems.
	Therefore, it is highly desirable to find an optimal network structure that is sufficiently small while the performance is kept reasonably good.
	
	During local critic training, each local critic network is trained to estimate the output of the main network eventually.
	Therefore, once the training of the proposed architecture finishes, we obtain different networks that are supposed to have similar input-output mappings but have different structures and possibly different accuracy, i.e., multiple sub-models and one main model (see Figure \ref{fig:structure_optimization}).
	Here, a sub-model is composed of the layers on the path from the input to a certain hidden layer and its local critic network.
	Among the sub-models, we can choose one as a structure-optimized network by considering the trade-off relationship between the complexity and performance.
	
	It is worth mentioning that our structural optimization approach can be performed instantly after training of the model, whereas many existing methods for structural optimization require iterative search processes, e.g., \cite{zoph17}.
	
	%%%%%%%%%%%%%%%%%%%%%%%%
	\subsection{Progressive inference}
	
	We propose another simple but effective way to utilize the sub-models obtained by the proposed approach for computational efficiency, which we call \textit{progressive inference}.
	Although small sub-models (e.g., sub-model 1) tend to show low accuracy, they would still perform well for some data.
	For such data, we do not need to perform the full feedforward pass but can take the classification decision by the sub-models.
	Thus, the basic idea of the progressive inference is to finish inference (i.e., classification) with a small sub-model if its confidence on the classification result is high enough, instead of completing the full feedforward pass with the main model, which can reduce the computational complexity.
	Here, the softmax outputs for all classes are compared and the maximum probability is used as the confidence level.
	If it is higher than a threshold, we take the decision by the sub-model; otherwise, the feedforward pass continues.
	The proposed progressive inference method is summarized in Algorithm 1\footnote{Our method shares some similarity with the anytime prediction scheme \cite{larsson17,huang18multi} that produces outputs according to the given computational budget. However, ours does not require particular network structures (such as multi-scale dense network \cite{huang18multi} or FractalNet \cite{larsson17}) but works with generic CNNs.}.
	
	\begin{algorithm}[t]
		\caption{Progressive inference}
		\label{alg_progressive}
		\begin{small}
		\begin{algorithmic}
			\STATE {\bfseries Input:} data $x$, threshold $t$
			\STATE {\bfseries Model:} sub-model $c_{i}$, main-model $f$
			\STATE Initialize: $classification = 0$.
			\FOR{$i=1$ {\bfseries to} $N-1$}
			\IF{$max$ softmax($c_{i}(x)$) $>$ $t$}
			\STATE $classification = argmax$ softmax($c_{i}(x)$) 
			\STATE $break$
			\ENDIF
			\ENDFOR
			\IF{$classification == 0$}
			\STATE \# if all sub-models are not confident
			\STATE $classification = argmax$ softmax($f(x)$)
			\ENDIF
		\end{algorithmic}
	 \end{small}
	\end{algorithm}
	
	%%%%%%%%%%%%%%%%%%%%%%%%
	\subsection{Ensemble inference}
	
	In recent deep learning systems, it is popular to use ensemble approaches to improve performance in comparison to single models, where multiple networks are combined for producing final results, e.g., \cite{he16deep,szegedy15}.
	The sub-models and  main model obtained by applying the proposed local critic training approach can be used for ensemble inference.
	Figure \ref{fig:ensemble} depicts how the sub-models and the main model can work together to form an ensemble classifier.
	We take the simplest way to combine them, i.e., summation of the networks' outputs.
	
	%%%%%%%%%%%%%%%%%%%%%%%%
	\section{Experiments}
	\label{sec:exp}
	
	\begin{figure*}[t!]
		\centering
		\begin{subfigure}[t]{0.49\textwidth}
			\centering
			\includegraphics[height=1.3in]{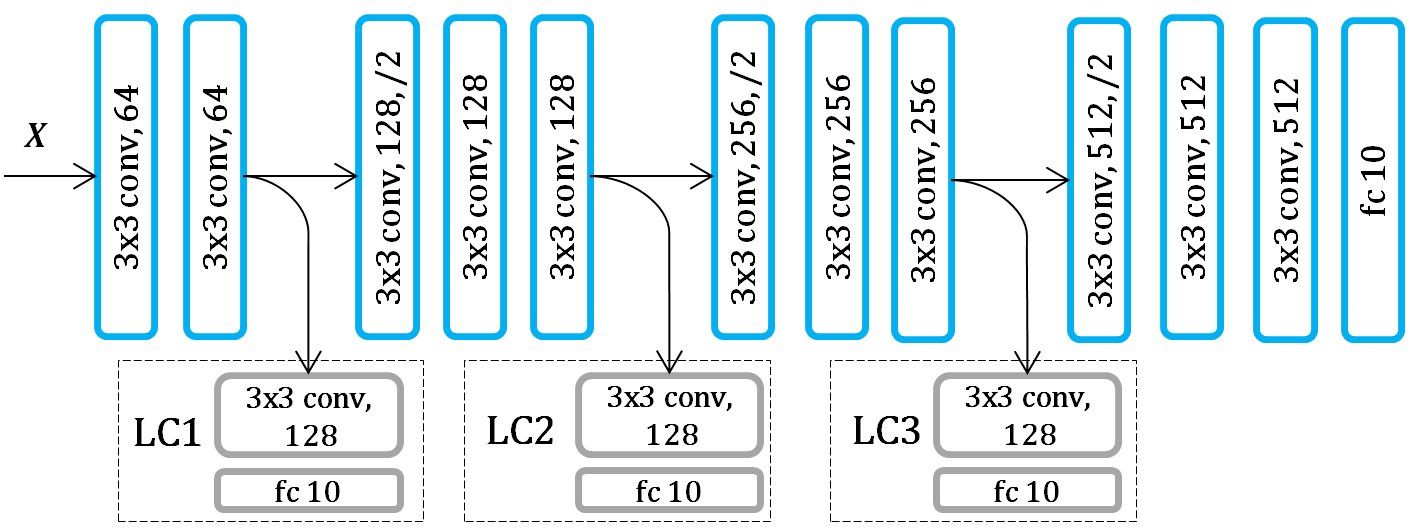}
			\caption{}\label{fig:experiment_model}%
		\end{subfigure}%
		~ 
		\begin{subfigure}[t]{0.49\textwidth}
			\centering
			\includegraphics[height=1.5in]{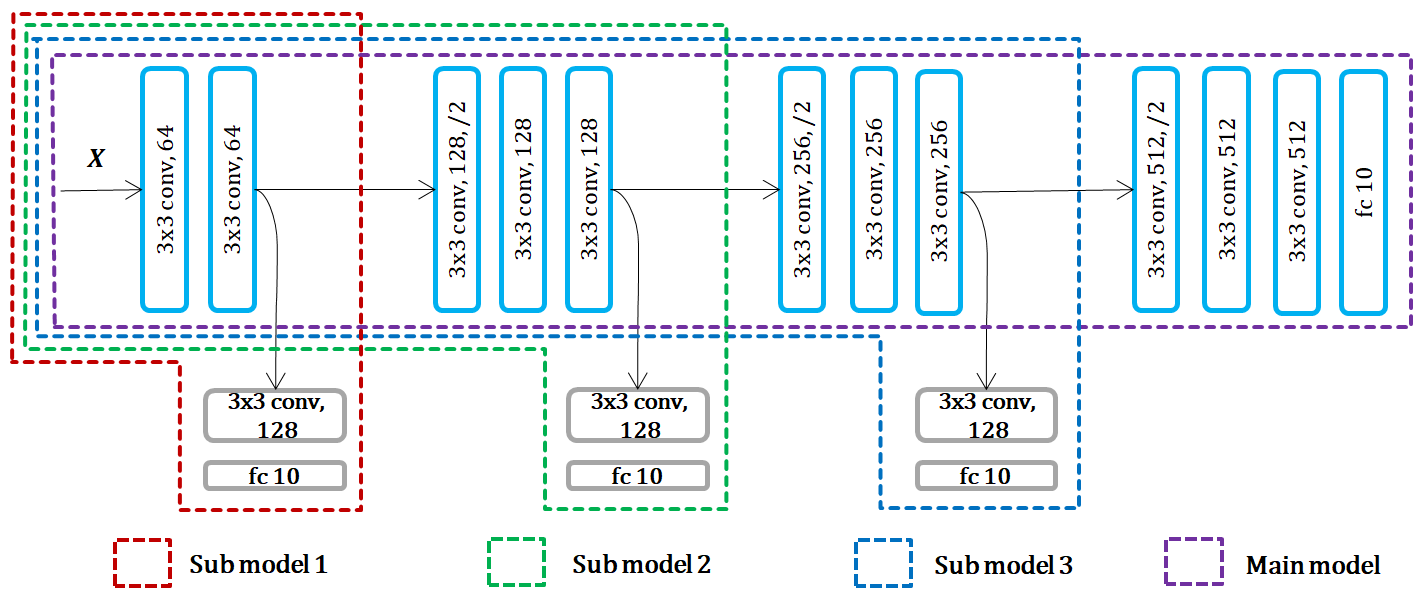}
			\caption{}\label{fig:structure_optimization}%
		\end{subfigure}
		\caption{(a) Network structure of the proposed approach using three local networks for CIFAR-10. LC1, LC2, and LC3 are local critic networks, each of which contains one convolutional layer. For CIFAR-100, the final fc10 layers of the main network and the local critic networks are replaced with fc100. (b) Sub-models and main model obtained by the proposed approach.}
	\end{figure*}
	
	We conduct extensive experiments to examine the performance of the proposed method in various aspects. 
	We use the CIFAR-10 and CIFAR-100 datasets \cite{Krizhevsky09} with data augmentation.
	We employ a VGG-like CNN architecture with batch normalization and ReLU activation functions, which is shown in Figure \ref{fig:experiment_model}.
	Note that this structure is the same to that used in \cite{Czarnecki17}. 
	It has three local critic networks, thus four layer groups that can be trained independently are formed (i.e., $N$=4).
	The local critic networks are also CNNs, and their structures are kept as simple as possible in order to minimize the computational complexity for computing the estimated error gradient given by (\ref{eq3}).
	
	We use the stochastic gradient descent with a momentum of 0.9 for the main network and the Adam optimization with a fixed learning rate of $10^{-4}$ for the local networks.
	The L2 regularization is used with $5\times 10^{-4}$ for the main network.
	For the loss functions in (\ref{eq2}) and (\ref{eq4}), the cross-entropy and the L1 loss are used, respectively, which is determined empirically. 
	The batch size is set to 128, and the maximum training iteration is set to 80,000.
	The learning rate for the main network is initialized to 0.1 and dropped by an order of magnitude after 40,000 and 60,000 iterations.
	The Xavier method is used for initialization of the network parameters.
	All experiments are performed using TensorFlow.
	We conduct all the experiments five times with different random seeds and report the average accuracy.
	
	\subsection{Performance evaluation}
	
	\begin{figure*}[t!]
		\centering
		\begin{subfigure}[t]{0.49\textwidth}
			\centering
			\includegraphics[width=0.98\textwidth, ]{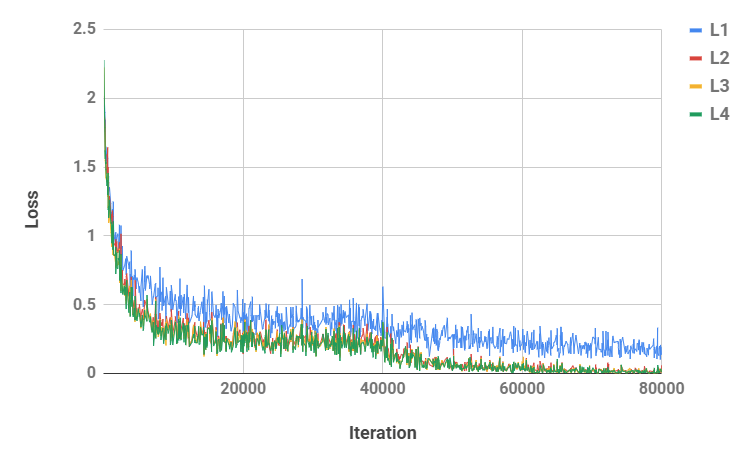}
			\caption{CIFAR-10}\label{fig:loss_cifar10}%
		\end{subfigure}%
		~ 
		\begin{subfigure}[t]{0.49\textwidth}
			\centering
			\includegraphics[width=0.98\textwidth, ]{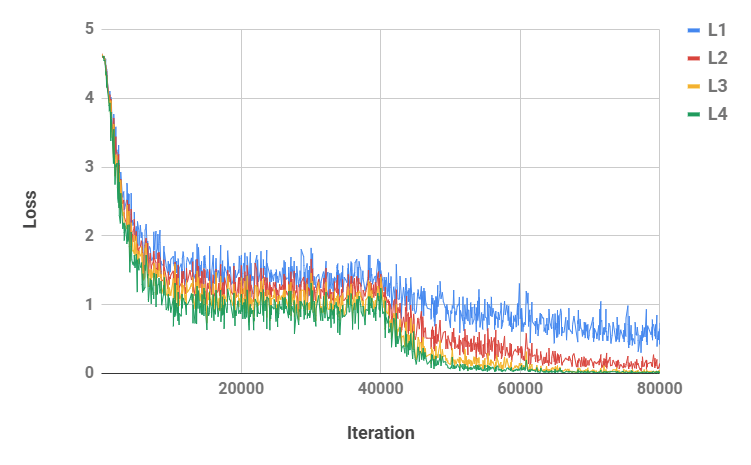}
			\caption{CIFAR-100}\label{fig:loss_cifar100}%
		\end{subfigure}
		\caption{Training loss values of the main model and each sub-model with respect to the training iteration.}
		\label{fig:loss}
	\end{figure*}
	
	\begin{table*}[t]
		\caption{Average test accuracy (\%) of backpropagation (BP), DNI \cite{Jaderberg17}, critic training \cite{Czarnecki17}, and proposed local critic training (LC). The numbers of local networks used are shown in the parentheses. The standard deviation values are also shown.}
		\label{table1}
		\centering
		\begin{tabular}{lccccccr}
			\toprule
			Dataset & BP & DNI (3) & Critic (3) & LC (1) & LC (3) & LC (5) \\
			\midrule
			CIFAR-10  & 93.93 \tiny{$\pm$0.20} & 64.86 \tiny{$\pm$0.42} & 91.92 \tiny{$\pm$0.30} & 92.06 \tiny{$\pm$0.20} & 92.39 \tiny{$\pm$0.09} & 91.38 \tiny{$\pm$0.20} \\
			CIFAR-100 & 75.14 \tiny{$\pm$0.18} & 36.53 \tiny{$\pm$0.64} & 69.07 \tiny{$\pm$0.25}  & 73.61 \tiny{$\pm$0.31} & 69.91 \tiny{$\pm$0.50} & 63.53 \tiny{$\pm$0.24} \\
			\bottomrule
		\end{tabular}
	\end{table*}

	%First, we validate the performance of the proposed local critic training approach. 
	Figure \ref{fig:loss} shows how the loss values of the main network and each local critic network, i.e., $L_i$ in (\ref{eq2}), evolve with respect to the training iteration.
	The graphs show that the local critic networks successfully learn to approximate the main network's loss with high accuracy during the whole training process.
	The local critic network farthest from the output side (i.e., $L_1$) shows larger loss values than the others, which is due to the inaccuracy accumulated through the cascaded approximation.  
	
	The classification performance of the proposed local critic training approach is evaluated in Table \ref{table1}. 
	For comparison, the performance of the regular backpropagation, DNI \cite{Jaderberg17}, and critic training \cite{Czarnecki17} is also evaluated. Although the critic training method is not for removing update locking, we include its result because it shares some similarity with our approach, i.e., additional modules to estimate the main network's output. In all three methods, each additional module is composed of a convolutional layer and an output layer.
	In the case of the proposed method, we test different numbers of local critic networks. Figure \ref{fig:experiment_model} shows the structure with three local critic networks. When only one local network is used, it is located at the place of LC2 in Figure \ref{fig:experiment_model}. When five local networks are used, they are placed after every two layers of the main network. 
	
	When compared to the result of backpropagation, the proposed approach successfully decouples training of the layer groups at a small expense of accuracy decrease (note that the performance of the proposed method can be made closer to that of backpropagation using different structures, as will be shown in Tables \ref{table2} and Figure \ref{table8}). The degradation of the accuracy of our method is larger for CIFAR-100, which implies that the influence of gradient estimation is larger for more complex problems. 
	When more local critic networks are used, the accuracy tends to decrease more due to higher reliance on predicted gradients rather than true gradients, while more layer groups can be trained independently. 
	%However, for CIFAR-10, LC (3) achieved higher performance than LC (1).
	%That is, it is important to use an appropriate number of LC according to the data.
	Thus, there exists a trade-off between the accuracy and unlocking effect.
	%Also, the proposed approach outperforms both DNI and critic training. 
	The DNI method shows poor performance as in \cite{Czarnecki17}.
	The proposed method shows similar (or even slightly better) performance to the critic training method, which shows the efficacy of the cascaded learning scheme of the local networks in our method. 

	\subsection{Structures of local critic networks}
	\begin{table*}[t]
		\caption{Average test accuracy (\%) with respect to the number of layers in the local critic networks. $[a,b,c]$ means that the numbers of convolutional layers in LC1, LC2, and LC3 are $a$, $b$, and $c$, respectively.}
		\label{table2}
		\centering
		\begin{scriptsize}
			\begin{tabular}{lccccccr}
				\toprule
				Dataset & [1,1,1] (default) & [3,3,3] & [5,5,5] & [3,2,1] &
				[1,2,3] & [5,4,3] & [3,4,5]\\
				\midrule
				CIFAR-10  & 92.39 \tiny{$\pm$0.09} & 92.36 \tiny{$\pm$0.22} & 91.72 \tiny{$\pm$0.19} & 92.07 \tiny{$\pm$0.21} & 92.20 \tiny{$\pm$0.12} & 92.10 \tiny{$\pm$0.16} & 91.90 \tiny{$\pm$0.16}\\
				CIFAR-100 & 69.91 \tiny{$\pm$0.50} &70.02 \tiny{$\pm$0.29} & 70.34 \tiny{$\pm$0.16} & 70.06 \tiny{$\pm$0.64} & 69.81 \tiny{$\pm$0.33} & 70.87 \tiny{$\pm$0.40} & 69.93 \tiny{$\pm$0.56}\\
				\bottomrule
			\end{tabular}
		\end{scriptsize}
	\end{table*}
	
	We examine the influence of the structures of the local critic networks in our method.
	Two aspects are considered, one about the influence of the overall complexity of the local networks and the other about the relative complexities of the local networks for good performance.
	For this, we change the number of convolutional layers in each local critic network, while keeping the other structural parameters unchanged.
	
	The results for various structure combinations of the three local critic networks are shown in Table \ref{table2}.
	As the number of convolutional layers increase for all local networks (the first three cases in the table),
	the accuracy for CIFAR-100 slightly increases from 69.91\% (with one convolutional layer) to 70.02\% (three convolutional layers) and 70.34\% (five convolutional layers), whereas for CIFAR-10 the accuracy slightly decreases when five convolutional layers are used.
	A more complex local network can learn better the target input-output relationship, which leads to the performance improvement for CIFAR-100.
	For CIFAR-10, on the other hand, the network structure with five convolutional layers seems too complex compared to the data to learn, which causes the performance drop.
	%Therefore, it is necessary to select the appropriate structure of local critic network considering the relationship between data and performance.

	%Also, when we make the structure more complex, the time for training also increases (roughly by 50\% and 100\% with three and five convolutional layers, respectively, in our experiment).
	%Therefore, the complexity of the local networks should be determined by considering the trade-off between the computational complexity and classification accuracy.
	
	Next, the numbers of layers of the local networks are adjusted differently in order to investigate which local networks should be more complex for good performance.
	The results are shown in the last four columns of Table \ref{table2}.
	%The configuration $[3,2,1]$ performs better than $[1,2,3]$ and $[5,4,3]$ shows better performance than $[3,4,5]$.
	Overall, it is more desirable to use more complex structures for the local networks closer to the input side of the main model.
	For instance, LC1 and LC3 are supposed to learn the relationship from $h_1$ to $h_4$ and that from $h_3$ to $h_4$, respectively.
	More layers are involved from $h_1$ to $h_4$ in the main network, so the mapping that LC1 should learn would be more complicated, requiring a network structure with sufficient modeling capability.
	
	\subsection{Periodic update of local critic networks}
	
	\begin{table*}[t]
		\caption{Average test accuracy (\%) with respect to the update frequency of local critic networks.}
		\label{table3}
		\centering
		\begin{tabular}{llllll}
			\toprule
			Dataset & 1/1 & 1/2 & 1/3 & 1/4 & 1/5 \\
			\midrule
			CIFAR-10  & 92.39 \tiny{$\pm$0.09} & 91.91 \tiny{$\pm$0.19} & 91.78 \tiny{$\pm$0.18} & 91.57 \tiny{$\pm$0.12} & 91.35 \tiny{$\pm$0.17} \\
			CIFAR-100 & 69.91 \tiny{$\pm$0.50} & 67.99 \tiny{$\pm$0.49} & 67.76\tiny{$\pm$0.19} & 66.74 \tiny{$\pm$0.41} & 66.39 \tiny{$\pm$0.39} \\
			\bottomrule
		\end{tabular}
	\end{table*}
	
	A way to increase the efficiency of the proposed approach is to update the local critic networks not at every iteration but only periodically.
	This may degrade the accuracy but has two benefits.
	First, the amount of computation required to update the local networks can be reduced.
	Second, the burden of the communication between the layer groups also can be reduced.
	These benefits will be more significant when the local networks have larger sizes.
	
    For the default structure shown in Figure \ref{fig:experiment_model}, we compare different update frequency in Table \ref{table3}.
	It is noticed that the accuracy only slightly decreases as the frequency decreases.
	When the update frequency is a half of that for the main network (i.e., 1/2), the accuracy drops by 0.48\% and 1.92\% for the two datasets, respectively.
	Then, the decrease of the accuracy is only 0.56\% for CIFAR-10 and 1.60\% for CIFAR-100 when the update frequency decreases from 1/2 to 1/5.

	\subsection{Structural optimization}
	
	Table \ref{table5} compares the performance of the sub-models, and Table \ref{table6} shows the complexities of the sub-models in terms of the amount of computation for a feedforward pass and the number of weight parameters.
	A larger network (e.g., sub-model 3) shows better performance than a smaller network (e.g., sub-model 1), which is reasonable due to the difference in learning capability with respect to the model size.
	The largest sub-model (sub-model 3) shows similar accuracy to the main model (92.29\% vs. 92.39\% for CIFAR-10 and 67.54\% vs. 69.91\% for CIFAR-100), while the complexity is significantly reduced.
	For CIFAR-10, the computational complexity in terms of the number of floating-point operations (FLOPs) and the memory complexity are reduced to only about 30\% (15.72 to 4.52 million FLOPs, and 7.87 to 2.26 million parameters), as shown in Table \ref{table6}.
	If an absolute accuracy reduction of 1.86\% (from 92.39\% to 90.53\%) is allowed by taking sub-model 2, the reduction of complexity is even more remarkable, up to about one ninth.
	%We expect that it may be also possible to have a case where the main model shows low accuracy due to overfitting but a sub-model obtained via local critic training performs well with high accuracy, which would need further studies.
	
	In addition, the table also shows the accuracy of the networks that have the same structures with the sub-models but are trained using regular backpropagation.
	Surprisingly, such networks do not easily reach accuracy comparable to that of the sub-models obtained by local critic training, particularly for smaller networks (e.g., 74.46\% vs. 85.24\% with sub-model 1 for CIFAR-10).
	We think that joint training of the sub-models in local critic training helps them to find better solutions than those reached by independent regular backpropagation.
	
	Therefore, these results demonstrate that a structurally optimized network can be obtained at a cost of a small loss in accuracy by local critic training, which may not be attainable by trial-and-error with backpropagation.
	
	\begin{table*}[t]
		\caption{Average test accuracy (\%) of the sub-models produced by local critic training and the networks trained by regular backpropagation.}
		\label{table5}
		\centering
		\begin{tabular}{lccccccr}
			\toprule
			Dataset & BP sub 1 & LC sub 1 & BP sub 2 & LC sub 2 & BP sub 3 & LC sub 3 \\
			\midrule
			CIFAR-10  & 74.46 \tiny{$\pm$0.91} & 85.24 \tiny{$\pm$0.49} & 88.03 \tiny{$\pm$0.87}
			& 90.53 \tiny{$\pm$0.15} & 92.05 \tiny{$\pm$0.24} & 92.29 \tiny{$\pm$0.09}\\
			CIFAR-100 & 47.58 \tiny{$\pm$1.10} & 55.39 \tiny{$\pm$0.57} & 61.79 \tiny{$\pm$0.92}
			& 63.62 \tiny{$\pm$0.31} & 67.81 \tiny{$\pm$0.22} & 67.54 \tiny{$\pm$0.70}\\
			\bottomrule
		\end{tabular}
		
	\end{table*}
	
	\begin{table}[t]
		\caption{FLOPs required for a feedforward pass and numbers of model parameters in the sub-models and main model for CIFAR-10. Note that sub-model 2 has less FLOPs and parameters than sub-model 1 due to the pooling operation in sub-model 2.}
		\label{table6}
		\centering
		\begin{tabular}{lcccr}
			\toprule
			model & FLOP & $\#$ of parameters  \\
			\midrule
			Sub-model 1 & 2.85M & 1.42M \\
			Sub-model 2 & 1.76M & 0.88M \\
			Sub-model 3 & 4.52M & 2.26M \\
			Main model & 15.72M & 7.87M \\
			\bottomrule
		\end{tabular}
	\end{table}
	
	\begin{table}[t]
		\caption{Average FLOPs and accuracy of progressive inference for test data of CIFAR-10 when the threshold is set to 0.9 or 0.95.}
		\label{table7}
		\centering
		\begin{tabular}{lcccr}
			\toprule
			& FLOP & Accuracy (\%) \\
			\midrule
			Progressive inference \tiny{(0.9)} & 2.90M & 91.18 \tiny{$\pm$0.10} \\
			Progressive inference \tiny{(0.95)} & 3.05M & 91.75 \tiny{$\pm$0.16} \\
			Main model & 15.72M & 92.39 \tiny{$\pm$0.09} \\ 
			\bottomrule
		\end{tabular}
	\end{table}
	
	\subsection{Progressive inference}
	
	We apply the progressive inference algorithm shown in Algorithm 1 to the trained default network for CIFAR-10 with the threshold set to 0.9 or 0.95.
	The results are shown in Table \ref{table7}.
	The feedforward pass ends at different sub-models for different test data, and the average FLOPs over all test data are shown.
	When the threshold is 0.9, with only a slight loss of accuracy (92.39\% to 91.18\%), the computational complexity is reduced significantly, which is only 18.45\% of that of the main model.
	When the threshold increases to 0.95, the accuracy loss becomes smaller (only 0.64\%), while the complexity reduction remains almost the same (19.40\% of the main model's complexity).
	
	%We simply use the maximum class probability as the confidence level for a classification result in order to show the feasibility of the proposed idea.
	%However, there may be more intelligent ways to measure confidence for improved classification performance, which could be a future work topic.
	
	\subsection{Ensemble inference}
	
\begin{figure*}[t!]
		\centering
		\begin{subfigure}[t]{0.8\textwidth}
			\centering
			\includegraphics[height=2in]{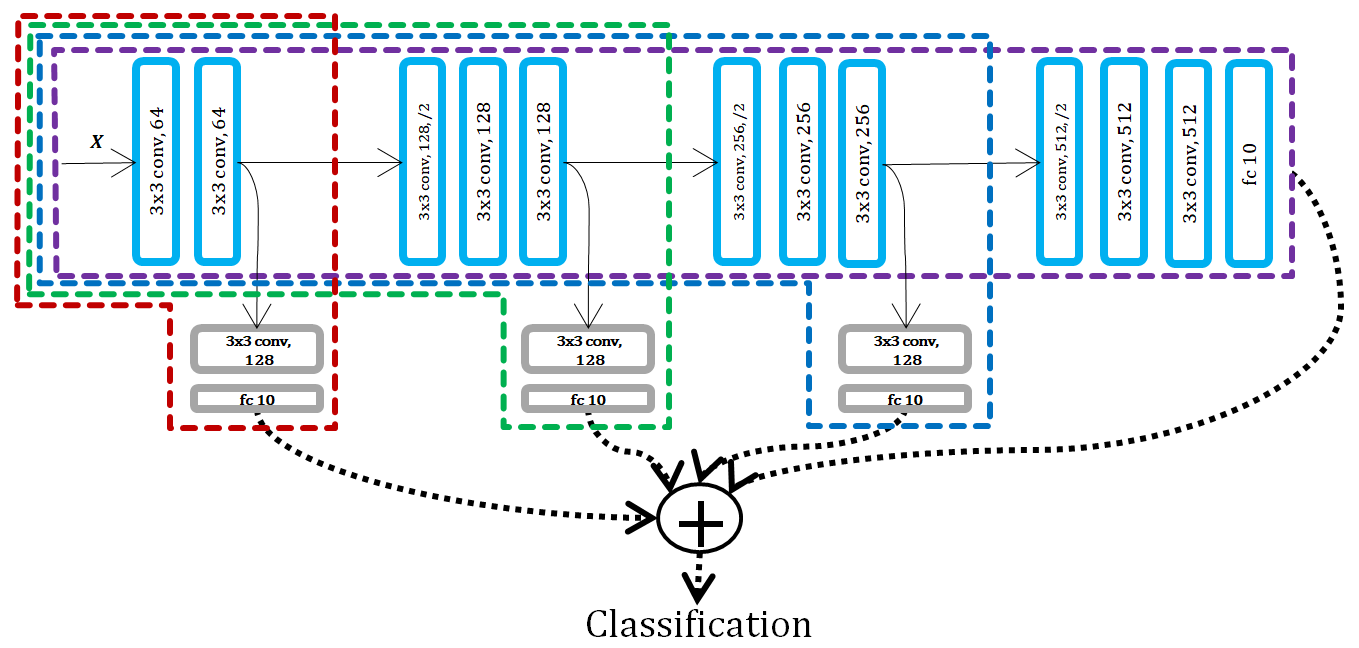}
			\caption{}\label{fig:ensemble}%
		\end{subfigure}%
		\\ %~ 
		\begin{subfigure}[t]{.9\textwidth}
			\begin{subfigure}[t]{0.5\textwidth}
				\centering
				\includegraphics[height=1.8in]{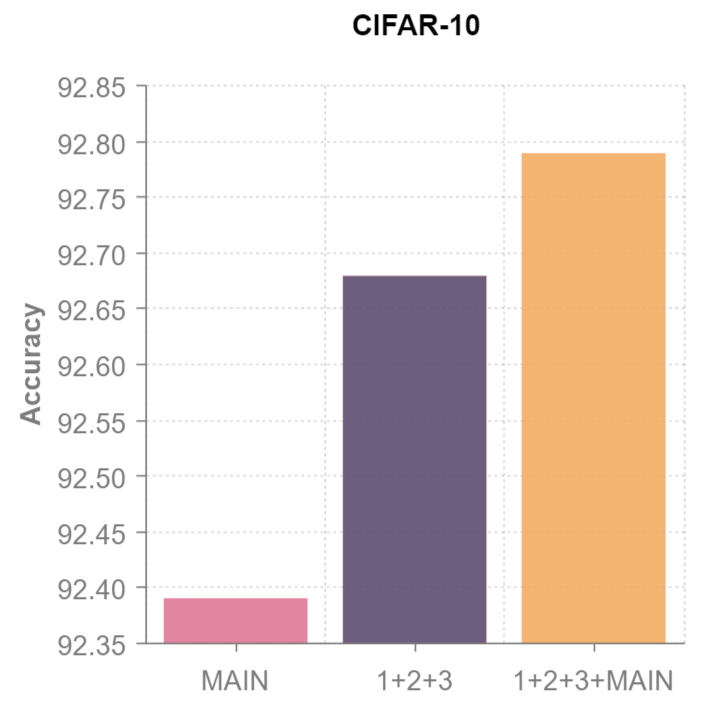}
			\end{subfigure}%
			~ 
			\begin{subfigure}[t]{0.5\textwidth}
				\centering
				\includegraphics[height=1.8in]{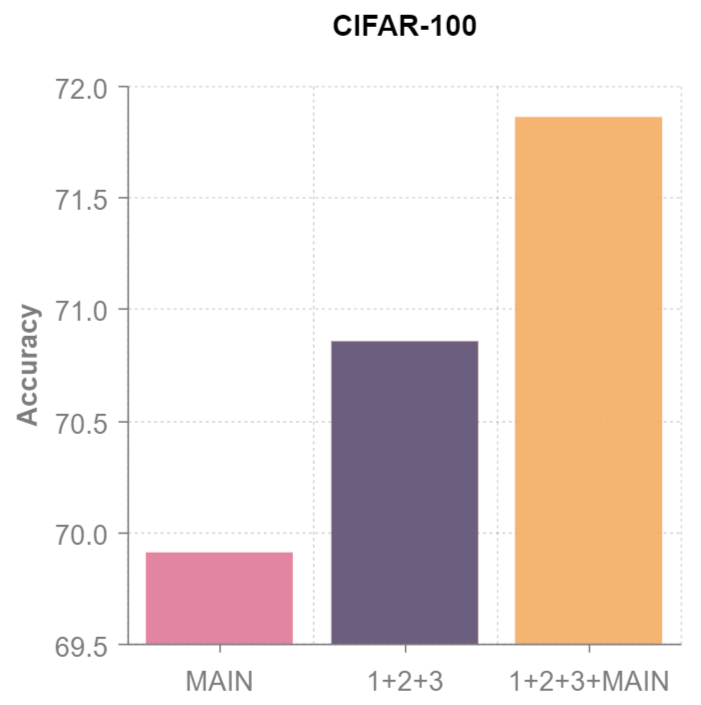}
			\end{subfigure}%
			\caption{}\label{table8}
		\end{subfigure}
		
		\caption{(a) Ensemble inference using the sub-models and main model. (b) Performance of the ensemble inference for an ensemble of the three sub-models (1+2+3) and an ensemble of the sub-models and the main model (1+2+3+main).}
	\end{figure*}

	The results of ensemble inference using the sub-models and main model are shown in Figure \ref{table8}.
	Using an ensemble of the three sub-models, we observe improved classification accuracy (92.68\% and 70.86\% for the two datasets, respectively) in comparison to the main model.
	The performance is further enhanced by an ensemble of both the three sub-models and the main model (92.79\% and 71.86\%).
	The improvement comes from the complementarity among the models, particularly between the models sharing a smaller number of layers.
	For instance, we found that sub-model 3 and the main model tend to show coincident classification results for a large portion of test data, so their complementarity is not significant;
	on the other hand, more data are classified differently by sub-model 1 and the main model, where we mainly observe performance improvement.
	Instead of the simple summation, there could be a better method to combine the models, which is left for future work.
	
	%%%%%%%%%%%%%%%%%%%%%%%%%%%%%%%%%%%%%%%%%%%%%%
	\section{Conclusion}
	\label{sec:conclusion}
	
	In this paper, we proposed the local critic training approach for removing the inter-layer locking constraint in training of deep neural networks. 
	In addition, we proposed three applications of the local critic training method: structural optimization of neural networks, progressive inference, and ensemble classification.
	It was demonstrated that the proposed method can successfully train CNNs with local critic networks having extremely simple structures.
	The performance of the method was also evaluated in various aspects, including effects of structures and update intervals of local critic networks and influences of the sizes of layer groups.
	%It was shown that there exist trade-offs between the number of local networks and the classification performance when these hyperparameters are to be adjusted. 
	Finally, it was shown that structural optimization, progressive inference, and ensemble classification can be performed directly using the models trained with the proposed approach without additional procedures.
	
	\bibliography{reference}

\begin{thebibliography}{10}

\bibitem{Carreira14}
M.~A. Carreira-Perpinan and W.~Wang.
\newblock Distributed optimization of deeply nested systems.
\newblock In {\em International Conference on Artificial Intelligence and
  Statistics (AISTATS)}, pages 10--19, Reykjavik, Iceland, 2014.

\bibitem{Cortes17}
C.~Cortes, X.~Gonzalvo, V.~Kuznetsov, M.~Mohri, and S.~Yang.
\newblock {AdaNet}: Adaptive structural learning of artificial neural networks.
\newblock In {\em International Conference on Machine Learning (ICML)}, pages
  874--883, Sydney, Australia, 2017.

\bibitem{Czarnecki17}
W.~M. Czarnecki, S.~Osindero, M.~Jaderberg, G.~Swirszcz, and R.~Pascanu.
\newblock Sobolev training for neural networks.
\newblock In {\em Advances in Neural Information Processing Systems (NIPS)},
  pages 4278--4287, Long Beach, CA, 2017.

\bibitem{feng15}
J.~Feng and T.~Darrell.
\newblock Learning the structure of deep convolutional networks.
\newblock In {\em International Conference on Computer Vision (ICCV)}, pages
  2749--2757, Santiago, Chile, 2015.

\bibitem{he16deep}
K.~He, X.~Zhang, S.~Ren, and J.~Sun.
\newblock Deep residual learning for image recognition.
\newblock In {\em IEEE Conference on Computer Vision and Pattern Recognition
  (CVPR)}, pages 770--778, Las Vegas, NV, 2016.

\bibitem{he16identity}
K.~He, X.~Zhang, S.~Ren, and J.~Sun.
\newblock Identity mappings in deep residual networks.
\newblock In {\em European Conference on Computer Vision (ECCV)}, pages
  630--645, Amsterdam, The Netherlands, 2016.

\bibitem{huang18multi}
G.~Huang, D.~Chen, T.~Li, F.~Wu, L.~Maaten, and K.Q. Weinberger.
\newblock Multi-scale dense networks for resource efficient image
  classification.
\newblock In {\em International Conference on Learning Representations (ICLR)},
  Vancouver, Canada, 2018.

\bibitem{Jaderberg17}
M.~Jaderberg, W.~M. Czarnecki, S.~Osindero, O.~Vinyals, A.~Graves, D.~Silver,
  and K.~Kavukcuoglu.
\newblock Decoupled neural interfaces using synthetic gradients.
\newblock In {\em International Conference on Machine Learning (ICML)}, pages
  1627--1635, Sydney, Australia, 2017.

\bibitem{Krizhevsky09}
A.~Krizhevsky.
\newblock {Learning multiple layers of features from tiny images}.
\newblock Master's thesis, Department of Computer Science, University of
  Toronto, 2009.

\bibitem{kwon97}
T.-Y. Kwok and D.-Y. Yeung.
\newblock Constructive algorithms for structure learning in feedforward neural
  networks for regression problems.
\newblock {\em IEEE Transactions on Neural Networks}, 8(3):630--645, 1997.

\bibitem{larsson17}
G.~Larsson, M.~Maire, and G.~Shakhnarovich.
\newblock {FractalNet}: Ultra-deep neural networks without residuals.
\newblock In {\em International Conference on Learning Representations (ICLR)},
  Toulon, France, 2017.

\bibitem{LeCun15}
Y.~LeCun, Y.~Bengio, and G.~Hinton.
\newblock Deep learning.
\newblock {\em Nature}, 521:436--444, 2015.

\bibitem{reed93}
R.~Reed.
\newblock Pruning algorithms- a survey.
\newblock {\em IEEE Transactions on Neural Networks}, 4(5):730--747, 1993.

\bibitem{szegedy15}
C.~Szegedy, W.~Liu, Y.~Jia, P.~Sermanet, S.~Reed, D.~Anguelov, D.~Erhan,
  V.~Vanhoucke, and A.~Rabinovich.
\newblock Going deeper with convolutions.
\newblock In {\em IEEE Conference on Computer Vision and Pattern Recognition
  (CVPR)}, pages 1--9, Boston, MA, 2015.

\bibitem{Taylor16}
G.~Taylor, R.~Burmeister, Z.~Xu, B.~Singh, A.~Patel, and T.~Goldstein.
\newblock Training neural networks without gradients: A scalable {ADMM}
  approach.
\newblock In {\em International Conference on Machine Learning (ICML)}, pages
  2722--2731, New York, NY, 2016.

\bibitem{zoph17}
B.~Zoph and Q.V. Le.
\newblock Neural architecture search with reinforcement learning.
\newblock In {\em International Conference on Learning Representations (ICLR)},
  Toulon, France, 2017.

\end{thebibliography}
	\bibliographystyle{plain}
	
\end{document}